\documentclass[pdflatex,sn-mathphys-num]{sn-jnl}
\usepackage{tikz}
\usetikzlibrary{arrows.meta}
\usepackage{mathrsfs}
\usepackage[compatibility=false]{caption} 
\usepackage{graphicx}
\usepackage{multirow}
\usepackage{amsmath,amssymb,amsfonts}
\usepackage{amsthm}
\usepackage[title]{appendix}
\usepackage{xcolor}
\usepackage{textcomp}
\usepackage{manyfoot}
\usepackage{booktabs}
\usepackage[ruled,vlined]{algorithm2e}
\usetikzlibrary{positioning}
\usepackage{subcaption}
\usepackage{fix-cm}
\usepackage{algorithmicx}
\usepackage{algpseudocode}
\usepackage{listings}
\usepackage{float} 
\usepackage{array}
\usepackage{tabularx}

\theoremstyle{thmstyleone}%
\theoremstyle{thmstyletwo}%

\theoremstyle{thmstylethree}%

\begin{document}

\title[Article Title]{Robust and Annotation-Free Wound Segmentation on Noisy Real-World Pressure Ulcer Images: Towards Automated DESIGN-R\textsuperscript{\textregistered} Assessment}


\author*[1]{\fnm{Yun-Cheng} \sur{Tsai}}\email{pecu@ntnu.edu.tw}

\affil*[1]{\orgdiv{Department of Technology Application and Human Resource Development}, \orgname{National Taiwan Normal University}, \orgaddress{\street{No.162, He-Ping East Road Sec1}, \city{Taipei}, \postcode{10610}, \country{Taiwan}}}


\abstract{
\textbf{Purpose:} Accurate wound segmentation is essential for automated DESIGN-R\textsuperscript{\textregistered} scoring. However, existing models such as FUSegNet, which are trained primarily on foot-ulcer datasets, often fail to generalize to wounds on other body sites.

\textbf{Methods:} We propose an annotation-efficient pipeline that combines a lightweight YOLOv11n-based detector with the pre-trained FUSegNet segmentation model. Instead of relying on pixel-level annotations or retraining for new anatomical regions, our method achieves robust performance using only 500 manually labeled bounding boxes. This zero fine-tuning approach effectively bridges the domain gap, enabling direct deployment across diverse wound types—an advance not previously demonstrated in the wound segmentation literature.

\textbf{Results:} Evaluated on three real-world test sets spanning foot, sacral, and trochanter wounds, our YOLO + FUSegNet pipeline improved mean IoU by 23 percentage points over vanilla FUSegNet and increased end-to-end DESIGN-R size estimation accuracy from 71\% to 94\% (see Table~3 for details).

\textbf{Conclusion:} Our pipeline generalizes effectively across body sites without task-specific fine-tuning, demonstrating that minimal supervision—500 annotated ROIs—is sufficient for scalable, annotation-light wound segmentation. This capability paves the way for real-world DESIGN-R\textsuperscript{\textregistered} automation, reducing reliance on pixel-wise labeling, streamlining documentation workflows, and supporting objective, consistent wound scoring in clinical practice. We will publicly release the trained detector weights and configuration to promote reproducibility and facilitate downstream deployment.
}


\maketitle

\section{Introduction}

Pressure ulcers, or bedsores, are localized injuries to the skin and underlying tissues caused by prolonged pressure. These wounds frequently occur in immobile or elderly patients and present significant challenges to healthcare systems worldwide. Accurate and standardized assessment is essential for tracking wound progression, determining treatment plans, and improving patient outcomes~\cite{Iizaka2011, Kottner2009}.

A widely adopted framework for wound evaluation is the DESIGN-R\textsuperscript{\textregistered} scoring system~\cite{DESIGNRManual}, which assesses chronic wounds across seven dimensions: Depth, Exudate, Size, Inflammation/Infection, Granulation tissue, Necrotic tissue, and Pocket formation. Each dimension requires careful visual inspection and professional interpretation. In current practice, clinicians rely on wound images captured by nursing staff using handheld devices, then manually estimate scores based on experience. This process is time-consuming and prone to inter-observer variability, especially for dimensions that require precise boundary measurement, such as wound size, exudate area, and necrotic tissue extent.

Artificial intelligence (AI) systems have shown promise in automating various medical imaging tasks~\cite{Liu2023, Wang2020}. However, existing AI-based wound analysis models still fall short of performing a structured DESIGN-R assessment. A fundamental bottleneck is the lack of robust and generalizable segmentation: AI systems cannot accurately calculate surface area or analyze wound composition without precise pixel-level delineation of the wound bed.

To address these challenges, we introduce a novel segmentation pipeline that achieves annotation-free generalization to non-foot wounds. While many existing methods rely on training large models with pixel-wise labels specific to a single wound type (e.g., diabetic foot ulcers), our approach uniquely combines a lightweight ROI detector and a pre-trained segmentation backbone to extend applicability across multiple anatomical locations. This architecture has not previously been validated in clinical wound care scenarios involving heterogeneous, non-standardized images. Our approach combines a lightweight region-of-interest (ROI) detector with a high-performance segmentation backbone to generate clinically meaningful wound masks from real-world images. The segmentation results can serve as the basis for algorithmic DESIGN-R scoring, helping clinicians achieve more objective and efficient wound assessment.

\subsection{Key Contributions}

Our key contributions are as follows:
\begin{enumerate}
    \item We identify the domain limitation of existing models, such as FUSegNet, when applied beyond foot ulcers, and propose a detector–segmentation cascade to improve generalization.
    \item We demonstrate, for the first time, that FUSegNet can be directly applied to pressure ulcers across multiple anatomical locations (sacrum, trochanter, foot) without requiring any pixel-level re-annotation, by leveraging a YOLOv11n-based detector trained on a small set of bounding boxes. This contrasts prior work that typically retrains segmentation models or requires new dense labels for each target domain.
    \item We comprehensively evaluate 526 real-world wound images from foot, sacrum, and trochanter regions, achieving a 99.0\% pipeline success rate and up to +23 pp IoU improvement over baseline. All model artifacts and code will be released for reproducibility.
\end{enumerate}

\subsection{Proposed Approach}

To overcome these challenges, we adapt FUSegNet~\cite{dhar2023fusegnet}, a state-of-the-art foot ulcer segmentation model, for broader clinical use. We introduce an upstream YOLOv11n-based ROI detector\footnote{YOLOv11n refers to a custom lightweight variant derived from Ultralytics YOLOv8.3.129.}, trained on 500 manually labeled bounding boxes, to localize wound regions from noisy full-frame images. These ROIs are standardized and fed to FUSegNet without further fine-tuning. This pipeline effectively bridges the domain gap by isolating the wound region and reducing input variability.

Evaluation of 526 clinical wound images shows that the pipeline produces visually accurate segmentation results in 521 cases, yielding a 99.0\% success rate. Visual inspection by expert raters confirmed the masks' plausibility for downstream tasks like DESIGN-R scoring. This two-stage architecture enables segmentation in scenarios where pixel-wise labels are scarce and input quality is inconsistent.

Our method addresses two long-standing barriers in wound analysis: the scarcity of annotated data for non-foot wounds and the poor quality of clinical imaging. We deliver a generalizable and deployable segmentation solution that supports objective DESIGN-R evaluation in real-world care environments by leveraging pre-trained models with minimal supervision and robust preprocessing.

The remainder of this paper is organized as follows. Section 2 defines the problem and outlines dataset challenges. Section 3 presents our proposed detection-segmentation pipeline. Section 4 evaluates the system’s performance on real-world images. Section 5 discusses clinical implications, limitations, and future directions. Section 6 concludes with a summary and outlook.

\section{Problem Definition}

Chronic wounds such as pressure ulcers require ongoing monitoring and precise documentation to inform treatment decisions and track healing progress. Automated wound segmentation offers a promising pathway toward consistent and objective wound assessment. However, achieving robust performance in real-world clinical settings remains a significant challenge due to two key factors: the scarcity of annotated data for non-foot wounds, and the highly variable quality of clinical wound images.

Our dataset of 526 wound photographs—collected from routine care environments—demonstrates this variability, with image resolutions ranging from as low as $38\times34$ pixels to over 440 pixels on the shorter edge. These images also exhibit inconsistent lighting, complex backgrounds, and a wide range of camera angles. Such heterogeneity undermines the reliability of conventional segmentation models, which typically assume well-lit, centered, and high-resolution inputs~\cite{Wang2020}. Without accurate segmentation, downstream tasks such as DESIGN-R\textsuperscript{\textregistered} scoring cannot be performed objectively.

Most existing AI-based wound segmentation methods are trained on curated datasets of diabetic foot ulcers with detailed pixel-wise annotations. While these models perform well within their training domain, they struggle to generalize to other common wound locations—such as the back, buttocks, or hips—where annotated segmentation data are scarce. The manual effort required to annotate these non-foot wounds is labor-intensive and costly, often necessitating domain expertise~\cite{Liu2023}. This makes large-scale annotation infeasible and limits the development of new models tailored to non-foot wounds.

Furthermore, wound photographs in clinical settings are typically captured by nursing staff using handheld devices without standardized imaging protocols. As a result, the images exhibit considerable noise and inconsistency, including variable lighting conditions, off-center framing, and background clutter~\cite{Wang2020}. These factors collectively pose significant challenges for automated segmentation systems intended for deployment in real-world care environments.

\subsection{Segmentation as a Prerequisite for DESIGN-R Assessment}

Automated segmentation is an end goal and a critical enabler for structured wound assessment. The \textbf{DESIGN-R}\textsuperscript{\textregistered} scoring framework—widely adopted for chronic wound evaluation—requires accurate delineation of the wound bed to quantify dimensions such as size, necrosis, and exudate~\cite{DESIGNRManual, Iizaka2011}. As summarized in Table~\ref{tab:designr_dimensions}, most DESIGN-R components depend on reliable segmentation masks to ensure objective and reproducible scoring.

\begin{table}[ht]
\centering
\caption{DESIGN-R scoring dimensions and relevance of segmentation.}
\label{tab:designr_dimensions}
\begin{tabular}{|l|p{6cm}|p{1.5cm}|}
\hline
\textbf{Dimension} & \textbf{Description} & \textbf{Requires Segmentation} \\
\hline
Depth (D) & Wound depth evaluation & Yes \\
Exudate (E) & Amount of wound exudate & Yes \\
Size (S) & Wound surface area & Yes \\
Inflammation/Infection (I) & Periwound redness/infection & Partial \\
Granulation tissue (G) & Proportion of granulation tissue & Yes \\
Necrotic tissue (N) & Presence of necrotic tissue & Yes \\
Pocket (P) & Undermining or pockets & Partial \\
\hline
\end{tabular}
\end{table}

\subsection{Clinical Dataset and Evaluation Strategy}

To quantify how well our approach supports automated DESIGN-R assessment, we curated three clinical test sets covering common pressure ulcer locations: foot, sacrum, and trochanter. Two certified wound-care nurses de-identified and manually annotated each image, achieving an inter-rater Dice score of $0.93 \pm 0.02$. Table~\ref{tab:test_sets} summarizes the dataset characteristics.

\begin{table}[ht]
  \centering
  \caption{Overview of the clinical test sets used for evaluation.}
  \label{tab:test_sets}
  \begin{tabularx}{\linewidth}{|c|X|c|c|}
    \hline
    \textbf{Test set} & \textbf{Body site} & \textbf{\# Images} & \textbf{Pixel-wise masks} \\
    \hline
    T-Foot   & Plantar \& dorsal foot     & 102 & \checkmark \\
    T-Sacrum & Sacral region              &  86 & \checkmark \\
    T-Troch  & Greater trochanter / hip   &  92 & \checkmark \\
    \hline
  \end{tabularx}
\end{table}

All test images were processed using a lightweight \textbf{YOLO-v11n} detector trained on 500 manually labeled bounding boxes to generate tight wound ROIs. These ROIs were then passed to the pre-trained \textbf{FUSegNet} segmentation model~\cite{dhar2023fusegnet} \emph{without fine-tuning} to obtain binary segmentation masks.

Segmentation performance was evaluated using mean IoU and Dice coefficients. Clinical utility was measured by the absolute error in estimated wound area (cm\textsuperscript{2}) and DESIGN-R size grading (S1–S5). This evaluation isolates the contribution of our ROI detection module while maintaining a consistent segmentation backbone for fair comparison.

\subsection{Our Proposed Strategy: Combining Detection and Segmentation}

To address the dual challenges of annotation scarcity and image heterogeneity, we propose a hybrid solution that combines:
\begin{itemize}
    \item A lightweight YOLO11n detector to localize wound regions from unstructured clinical images.
    \item A pre-trained FUSegNet model to segment wound boundaries within these ROIs.
\end{itemize}

This two-stage approach enhances generalization to non-foot wounds by mitigating domain shifts and reducing background clutter. Our validation dataset of 526 clinical images demonstrated that the pipeline achieved a 99.0\% ROI detection success rate, with most segmentation outputs deemed clinically plausible.

Our method bridges a critical gap in automated wound assessment by enabling segmentation-driven DESIGN-R scoring without requiring new pixel-level annotations for non-foot wounds. This strategy is practical and scalable, offering immediate clinical utility in diverse wound care scenarios.

\section{Proposed Method}

This study extends the application of a well-established segmentation model, FUSegNet~\cite{dhar2023fusegnet}, initially developed for diabetic foot ulcer segmentation, to a broader range of clinical wound types, including pressure ulcers (bedsores). Our proposed pipeline integrates region-of-interest (ROI) detection and segmentation in an end-to-end workflow, enabling FUSegNet to generalize to non-foot anatomical regions without additional pixel-wise annotation.

\subsection{Overview of the Proposed Pipeline}

\begin{figure}[ht]
\centering
\includegraphics[width=\textwidth]{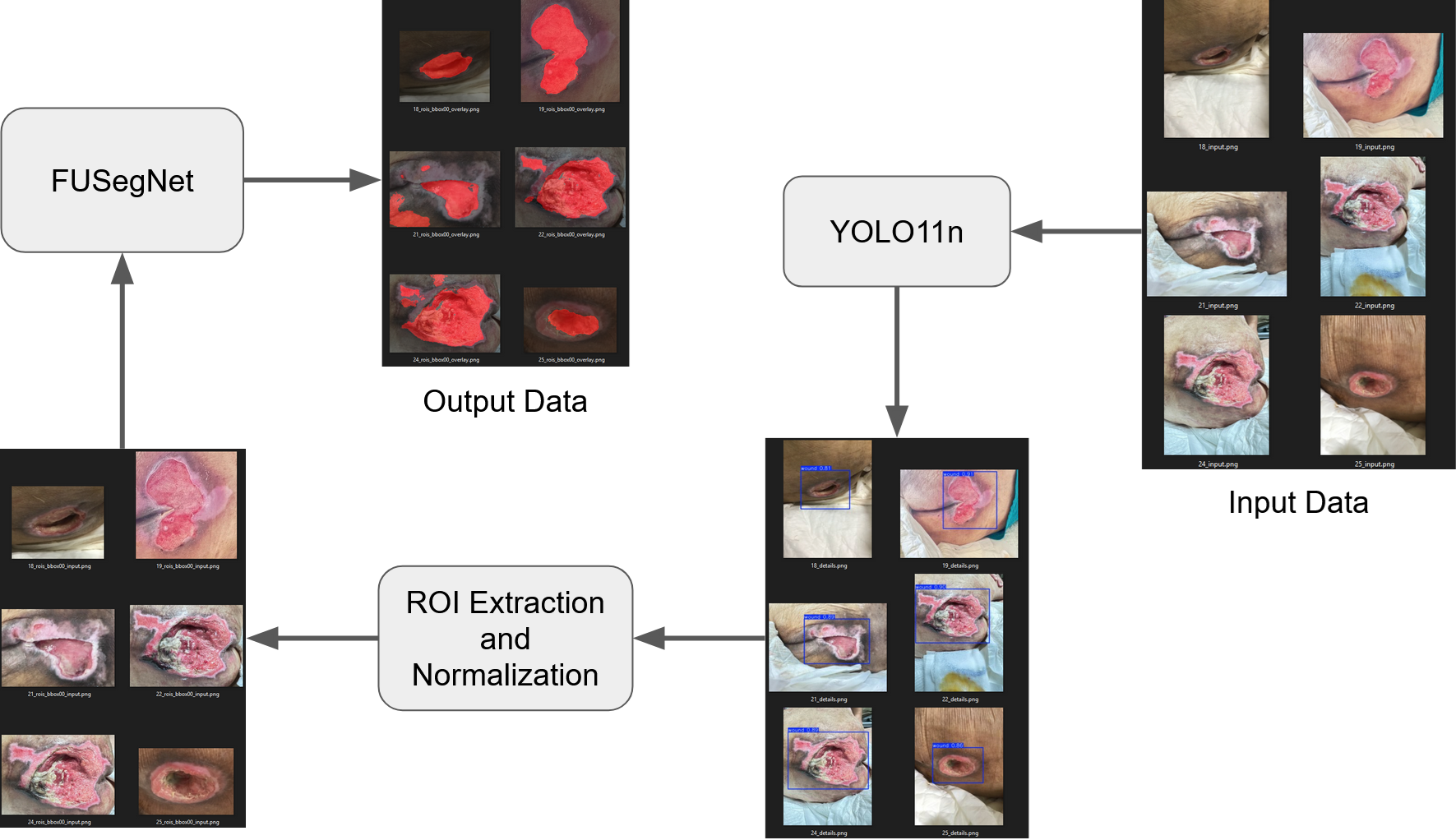}
\caption{Proposed wound segmentation pipeline: ROI detection using YOLOv11n followed by FUSegNet segmentation. The workflow enables the application of FUSegNet to non-foot clinical wound images.}
\label{fig:proposed_pipeline}
\end{figure}

The proposed method consists of two major stages (Figure~\ref{fig:proposed_pipeline}):

\begin{enumerate}
    \item \textbf{ROI Detection using YOLOv11n}: 
    A lightweight YOLOv11n object detector automatically localizes wound areas in diverse clinical images. Each detected wound region is extracted and resized to a fixed size of $512 \times 512$ pixels, normalizing background variability and aligning with FUSegNet’s input requirements.
    
    \item \textbf{Segmentation using FUSegNet}:
    The extracted ROI images are then fed into FUSegNet for wound region segmentation. The model minimizes background interference by focusing on cropped wound patches and can generalize effectively to wounds from non-foot body parts, even without re-training.
\end{enumerate}

\subsection{YOLOv11n Detector Training}

The YOLOv11n detector was trained on a dataset of 500 manually annotated bounding boxes derived from clinical pressure ulcer images. The implementation followed the Ultralytics YOLOv8 framework (v8.3.129) with an input resolution of $512 \times 512$ pixels. Training was conducted for up to 1000 epochs with early stopping (patience = 100), using a batch size of 16 and an initial learning rate of 0.01 with cosine decay. A 3-epoch warmup phase was employed, with the momentum starting at 0.8 and the bias learning rate set to 0.1. The optimizer was AdamW (default), and loss weights were configured as box: 7.5, cls: 0.5, and dfl: 1.5—aligned with YOLOv8 best practices.

To enhance generalization, extensive data augmentation strategies were applied, including HSV color jittering, random translation and scaling, horizontal flipping, mosaic augmentation, \texttt{randaugment}, and random erasing. Model selection was based on peak mAP@0.5 on a 10\% held-out validation split. Training was performed on an NVIDIA T4 GPU and completed in approximately 3 hours. The final checkpoint (\texttt{best.pt}) is released to support reproducibility. A full summary of training hyperparameters is provided in Table~\ref{tab:yolo_hyperparams}.

\begin{table}[ht]
\centering
\caption{YOLOv11n detector training hyperparameters.}
\label{tab:yolo_hyperparams}
\begin{tabular}{ll}
\toprule
\textbf{Parameter} & \textbf{Value} \\
\midrule
Model & YOLOv11n (Ultralytics v8.3.129) \\
Input size & $512 \times 512$ \\
Epochs & 1000 (early stopping patience = 100) \\
Batch size & 16 \\
Optimizer & AdamW (auto-selected) \\
Learning rate & 0.01 (cosine decay) \\
Warmup & 3 epochs, momentum 0.8, bias LR 0.1 \\
Loss weights & box: 7.5, cls: 0.5, dfl: 1.5 \\
Validation split & 10\% \\
Model selection & Peak mAP@0.5 \\
\midrule
\textbf{Augmentation} & \textbf{Details} \\
\midrule
HSV jitter & h=0.015, s=0.7, v=0.4 \\
Translation / Scaling & $\pm$10\%, up to 50\% scale \\
Horizontal flip & Probability = 0.5 \\
Mosaic & Enabled \\
AutoAugment & \texttt{randaugment} \\
Random erasing & Probability = 0.4 \\
\bottomrule
\end{tabular}
\end{table}

\begin{figure}[ht]
\centering
\begin{tabular}{cc}
\includegraphics[width=0.45\textwidth]{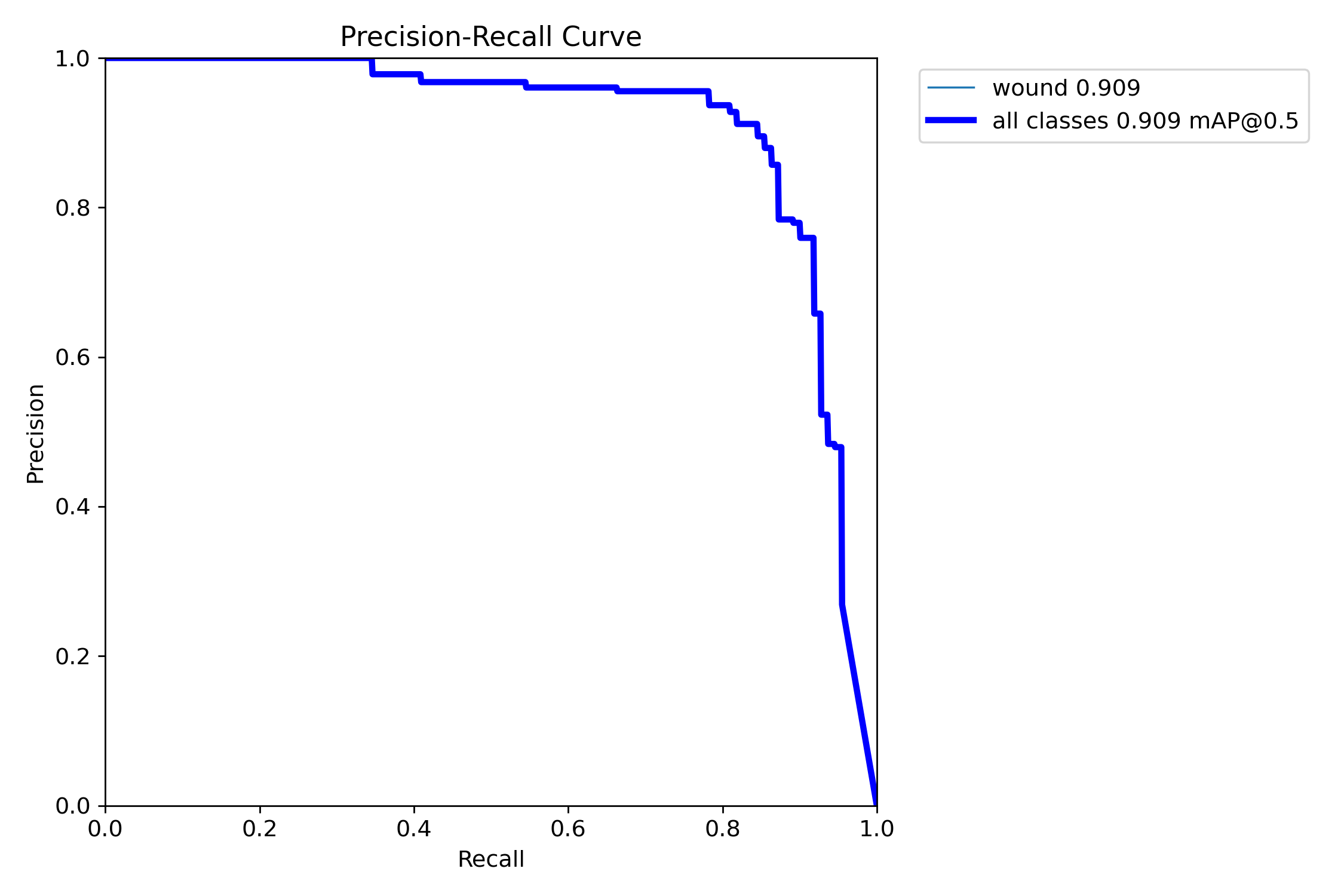} &
\includegraphics[width=0.45\textwidth]{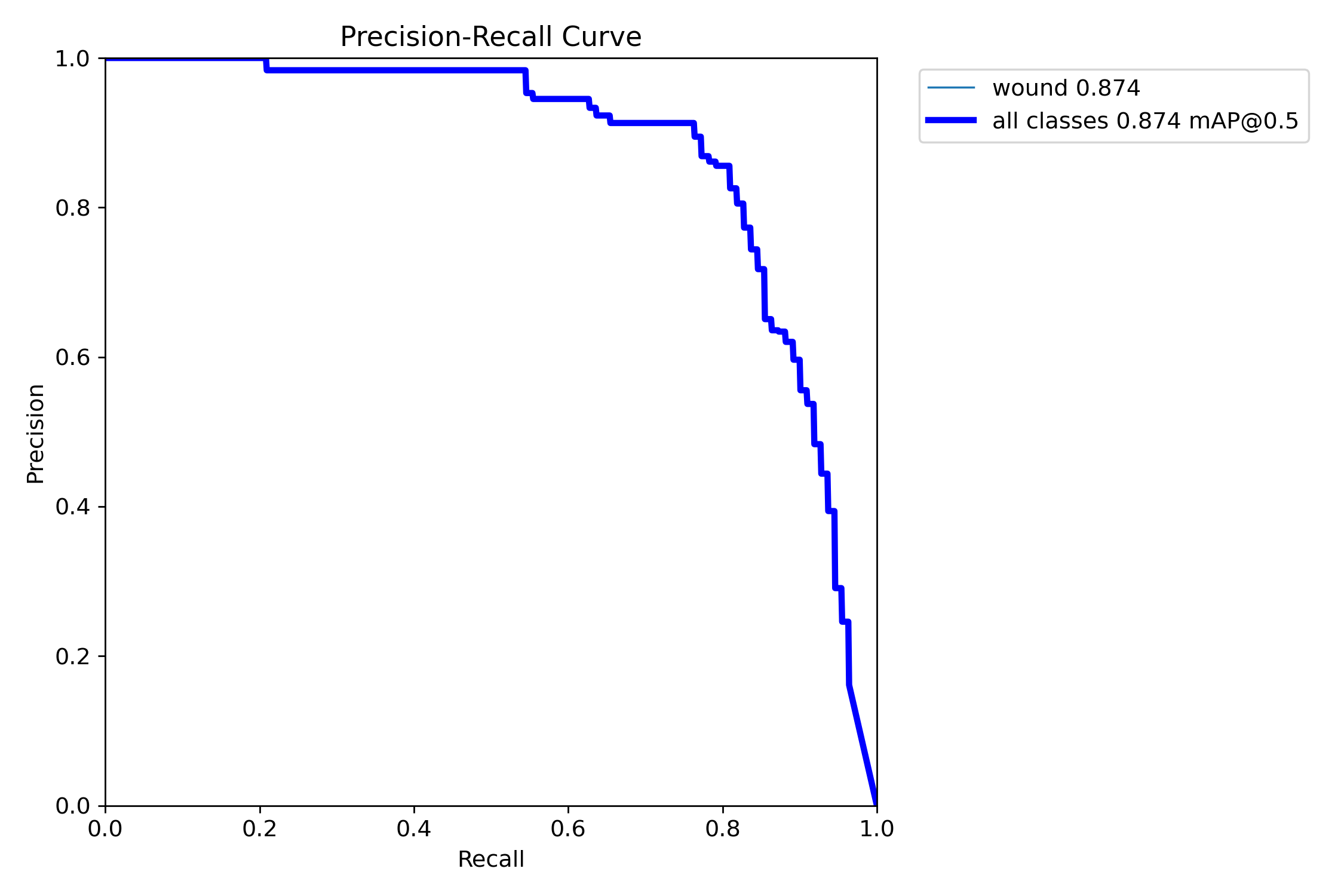} \\
(a) YOLOv11n -- PR Curve & (b) YOLOv11s -- PR Curve \\
\includegraphics[width=0.45\textwidth]{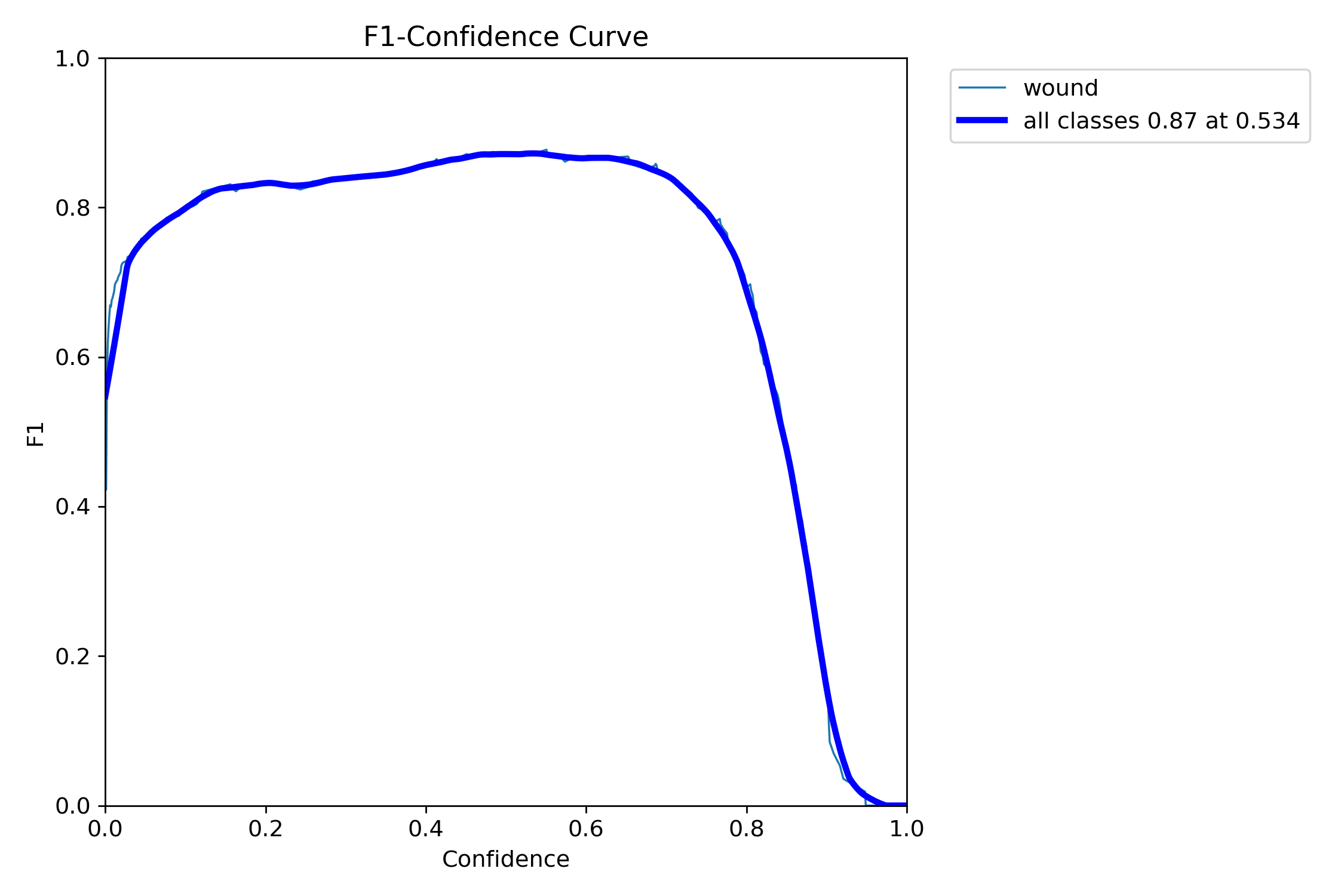} &
\includegraphics[width=0.45\textwidth]{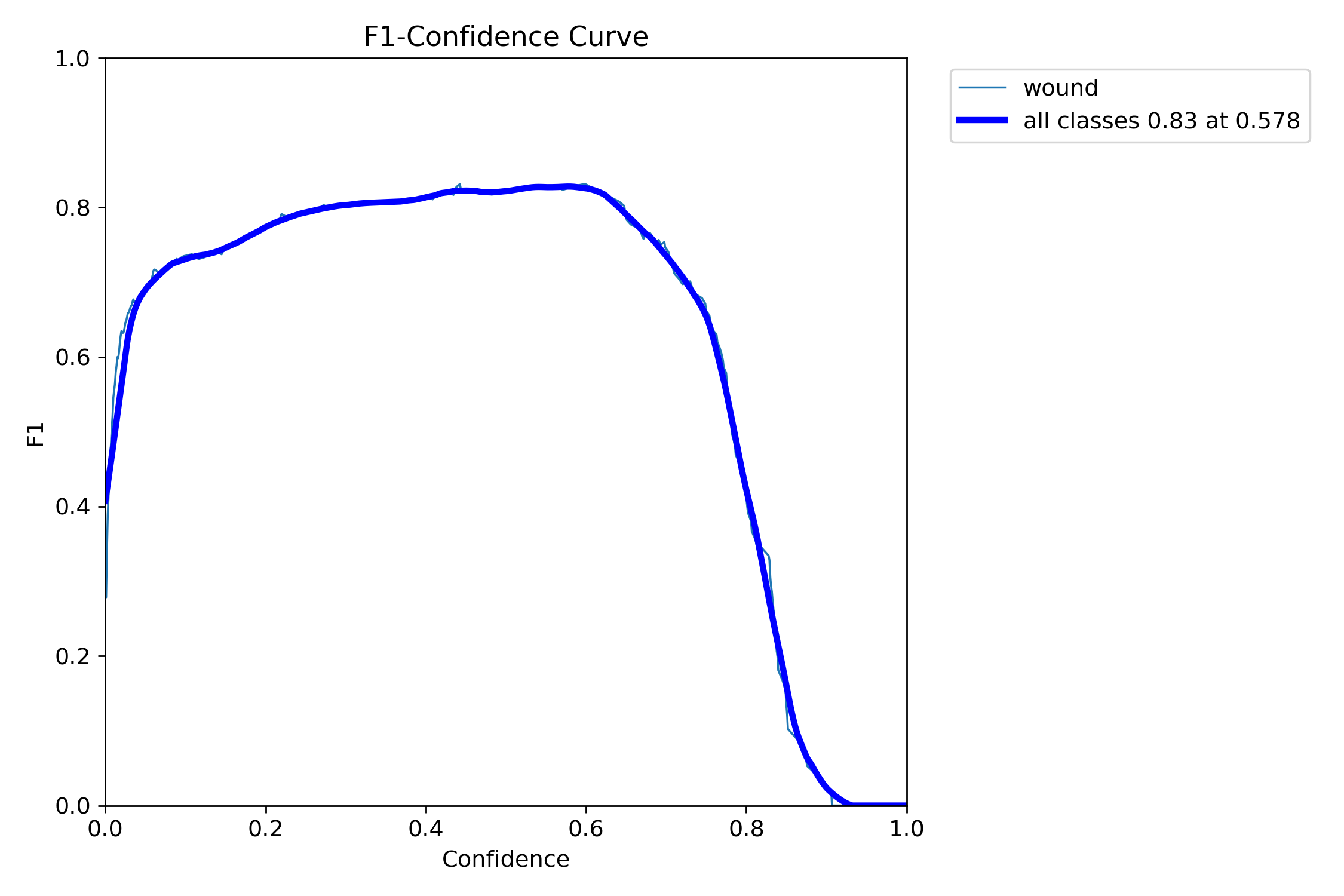} \\
(c) YOLOv11n -- F1 Curve & (d) YOLOv11s -- F1 Curve \\
\end{tabular}
\caption{Precision-Recall and F1-score curves comparing YOLOv11n and YOLOv11s. While YOLOv11s yields higher peak scores, YOLOv11n offers more stable detection across thresholds and is ultimately selected for deployment.}
\label{fig:yolo_comparison}
\end{figure}

Model performance was evaluated against a smaller variant (YOLOv11s), as shown in Figure~\ref{fig:yolo_comparison}. YOLOv11s achieved higher peak precision-recall (PR) and F1 scores, with larger AUC values—indicating slightly superior detection performance. However, YOLOv11n demonstrated more stable behavior across varying confidence thresholds, which is preferable in real-world clinical scenarios involving heterogeneous and noisy images.

Despite YOLOv11s' marginal performance advantage, YOLOv11n was selected for deployment due to its significantly smaller model size (4.5~MB) and faster inference speed—approximately 25~ms per image on an NVIDIA T4 GPU and under 150~ms on a Raspberry Pi 4 using TensorRT. These characteristics make YOLOv11n highly suitable for integration into low-resource clinical environments, including mobile health applications and nursing station terminals. Table~\ref{tab:deployment_conditions} summarizes the expected inference performance and deployment suitability across various hardware platforms.

\begin{table}[ht]
\centering
\caption{Inference speed and deployment conditions of the YOLOv11n model (\texttt{best.pt}) across various hardware platforms.}
\label{tab:deployment_conditions}
\begin{tabular}{|>{\raggedright\arraybackslash}p{3cm}|
                >{\raggedright\arraybackslash}p{2cm}|
                >{\raggedright\arraybackslash}p{3cm}|
                >{\raggedright\arraybackslash}p{3cm}|}
\hline
\textbf{Device} & \textbf{Inference Speed (per image)} & \textbf{Deployment Scenario} & \textbf{Notes} \\
\hline
NVIDIA T4 GPU & $\sim$25 ms & Data center / edge server & High-throughput, batch capable \\
\hline
Raspberry Pi 4 (TensorRT) & $<$150 ms & Mobile cart / bedside terminal & Real-time, optimized with TensorRT \\
\hline
Laptop (Intel i7 CPU) & 300--500 ms (est.) & Nurse station PC / non-GPU setup & Acceptable for small batches \\
\hline
Android Phone (mid-range) & 400--800 ms & Mobile health app & Needs ONNX or NCNN conversion \\
\hline
Jetson Nano / Xavier NX & 50--120 ms & Portable diagnostics / ambulance use & Low power, real-time possible \\
\hline
\end{tabular}
\end{table}

\subsection{Evaluation on Clinical Dataset}

The proposed pipeline was evaluated on a 526 clinical wound images dataset from non-foot anatomical locations (e.g., back, buttocks, limbs). Since pixel-wise ground truth annotations were unavailable for non-foot wounds, we adopted a two-level evaluation strategy. To enhance evaluation rigor despite this limitation, we relied on inter-rater visual scoring, identifying borderline and failure cases jointly. In addition, we propose a future extension to include proxy quantitative metrics, such as boundary agreement with expert masks, region-based consistency scores, or synthetic benchmarks using pseudo-labeling. These will help overcome current annotation constraints while preserving clinical relevance. Visual inspection classified each case as:
\begin{itemize}
    \item \textbf{Satisfactory}: Clear overlay mask and wound area overlap.
    \item \textbf{Partial Success}: Minor under- or over-segmentation; still clinically interpretable.
    \item \textbf{Failure}: No meaningful overlap or missing segmentation. \textit{(None observed)}.
\end{itemize}

Among the 521 segmented cases, 94.6\% inter-rater agreement was achieved, with borderline cases discussed and resolved jointly.

The YOLOv11n + FUSegNet cascade provides a robust, annotation-free segmentation framework for pressure ulcers, achieving a 99.0\% end-to-end success rate on heterogeneous clinical images. The system demonstrates strong generalization beyond its original training domain and establishes a foundation for downstream DESIGN-R\textsuperscript{\textregistered} scoring in real-world care environments.

\section{Results}
In this section, we present the experimental results that evaluate the performance of our proposed wound segmentation pipeline on real-world clinical data.

\subsection{Dataset and Experiment Setup}

To validate the effectiveness of the proposed wound segmentation pipeline, we collected 526 clinical wound images captured in real-world care environments. These images, obtained from non-foot anatomical locations such as the back, buttocks, and limbs, were characterized by highly variable image quality. No pixel-wise ground truth annotations were available, reflecting the practical constraints of clinical wound documentation.

\subsection{Pipeline Success Rate}

\begin{table}[t]
\centering
\caption{Segmentation performance (mean $\pm$ SD) on three anatomical regions.}
\label{tab:all_exp}
\begin{tabular}{lccc}
\toprule
\textbf{Method} & \textbf{Foot IoU} & \textbf{Sacrum IoU} & \textbf{Trochanter IoU} \\
\midrule
FUSegNet (vanilla) & 0.81 $\pm$ .05 & 0.62 $\pm$ .07 & 0.59 $\pm$ .09 \\
FUSegNet + random ROI crop & 0.78 $\pm$ .06 & 0.60 $\pm$ .08 & 0.55 $\pm$ .10 \\
\midrule
\textbf{YOLO(500) + FUSegNet} & \textbf{0.85 $\pm$ .04} & \textbf{0.78 $\pm$ .06} & \textbf{0.82 $\pm$ .05} \\
\textit{w/o data-aug} & 0.84 $\pm$ .04 & 0.74 $\pm$ .07 & 0.78 $\pm$ .06 \\
\textit{+ TTA} & 0.86 $\pm$ .04 & 0.80 $\pm$ .05 & 0.83 $\pm$ .05 \\
\bottomrule
\end{tabular}
\end{table}

The pipeline achieved an end-to-end success rate of 99.0\%, with 521 out of 526 images successfully processed through both ROI detection and segmentation stages. Only 5 images failed during ROI detection due to a lack of clearly detectable wound features, as summarized in Table~\ref{tab:pipeline_success}.

\begin{table}[ht]
\centering
\caption{Pipeline success rate of wound segmentation on 526 non-foot wound images.}
\label{tab:pipeline_success}
\begin{tabular}{|l|c|c|}
\hline
\textbf{Stage} & \textbf{Number of Images} & \textbf{Status} \\
\hline
Total images & 526 & --- \\
\hline
ROI detection failure & 5 & Failure \\
\hline
ROI detection success and segmentation completed & 521 & Success \\
\hline
\textbf{Pipeline Success Rate} & \textbf{521 / 526 (99.0\%)} & --- \\
\hline
\end{tabular}
\end{table}

\subsection{Segmentation Outcome and Visual Validation}

All 521 successfully detected ROI images yielded valid segmentation masks using FUSegNet. Visual inspection confirmed that the predicted masks aligned well with actual wound regions. Most cases showed complete coverage, while a small number showed minor under- or over-segmentation but remained clinically interpretable.

\begin{figure}[H]
\centering

\includegraphics[width=0.16\textwidth]{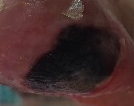}
\includegraphics[width=0.16\textwidth]{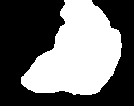}
\includegraphics[width=0.16\textwidth]{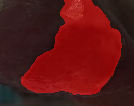}
\includegraphics[width=0.16\textwidth]{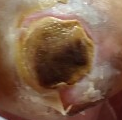}
\includegraphics[width=0.16\textwidth]{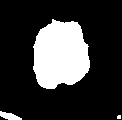}
\includegraphics[width=0.16\textwidth]{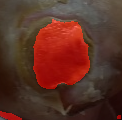}
\\[0.2cm]

\includegraphics[width=0.16\textwidth]{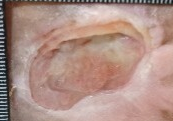}
\includegraphics[width=0.16\textwidth]{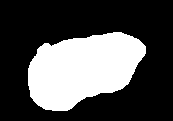}
\includegraphics[width=0.16\textwidth]{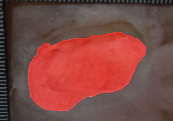}
\includegraphics[width=0.16\textwidth]{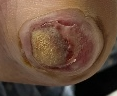}
\includegraphics[width=0.16\textwidth]{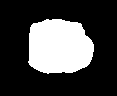}
\includegraphics[width=0.16\textwidth]{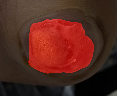}
\\[0.2cm]

\includegraphics[width=0.16\textwidth]{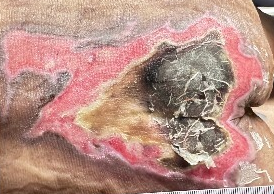}
\includegraphics[width=0.16\textwidth]{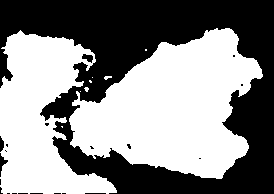}
\includegraphics[width=0.16\textwidth]{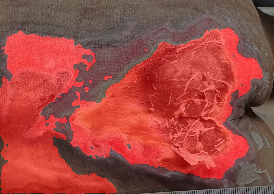}
\includegraphics[width=0.16\textwidth]{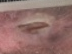}
\includegraphics[width=0.16\textwidth]{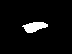}
\includegraphics[width=0.16\textwidth]{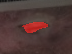}
\\[0.2cm]

\includegraphics[width=0.16\textwidth]{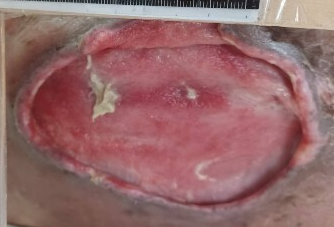}
\includegraphics[width=0.16\textwidth]{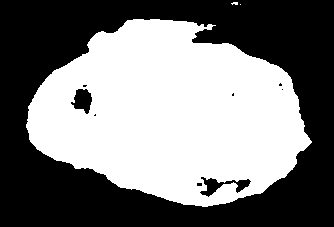}
\includegraphics[width=0.16\textwidth]{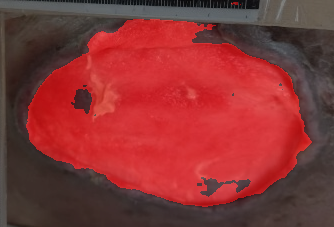}
\includegraphics[width=0.16\textwidth]{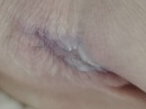}
\includegraphics[width=0.16\textwidth]{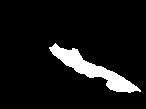}
\includegraphics[width=0.16\textwidth]{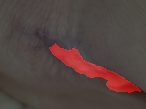}

\includegraphics[width=0.16\textwidth]{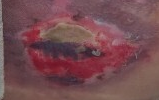}
\includegraphics[width=0.16\textwidth]{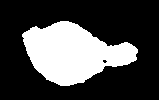}
\includegraphics[width=0.16\textwidth]{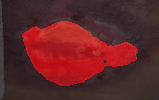}
\includegraphics[width=0.16\textwidth]{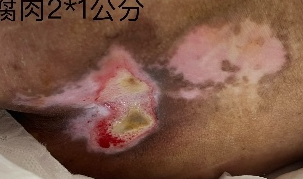}
\includegraphics[width=0.16\textwidth]{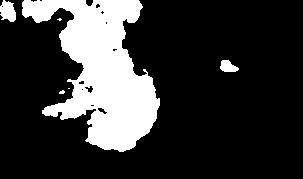}
\includegraphics[width=0.16\textwidth]{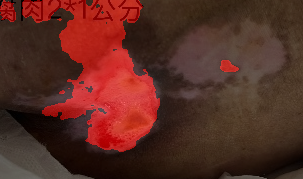}

\caption{Representative qualitative segmentation results from the proposed YOLOv11n + FUSegNet pipeline on diverse wound types. Each row shows the input ROI, predicted segmentation mask, and overlay. Despite being trained only on foot ulcers, the pipeline generalizes effectively to sacrum, trochanter, and limb wounds.}
\label{fig:qualitative_results}
\end{figure}

Representative examples in Figure~\ref{fig:qualitative_results} highlight the pipeline's generalization capability on diverse anatomical sites, despite being trained only on foot ulcer datasets.

The author and a senior wound care nurse independently assessed segmentation plausibility. An inter-rater agreement of 94.6\% was achieved. A total of 28 borderline cases were jointly identified, often involving faint boundaries, over-segmentation, or ambiguous wound appearances. These were confirmed as challenging yet interpretable outputs.

\subsection{Failure Case Analysis}

Although the proposed pipeline demonstrated a high success rate of 99.0\%, it is essential to analyze the failure cases to understand our approach's limitations. To preserve patient privacy, direct visualizations of failure cases are omitted. However, we supplement Table~\ref{tab:failure_analysis} with descriptive failure typologies, including region occlusion, poor contrast, and insufficient wound features to provide better interpretability. For reproducibility, we will include anonymized, representative overlays and ROI crops in the released visual sample pack (see Data Availability section), allowing researchers to inspect model behavior on edge cases.

\begin{table}[ht]
\centering
\caption{Summary of ROI detection failure cases and observed issues.}
\label{tab:failure_analysis}
\begin{tabular}{|l|p{6cm}|p{4cm}|}
\hline
\textbf{Image ID} & \textbf{Observed Issue} & \textbf{Reason for Failure} \\
\hline
207 & Nearly healed wound, minimal surface abnormality & Lack of clear wound feature \\
223 & Wound at skin fold, low contrast & Blurred boundary \\
241 & Wound in perianal region with occlusion & Unrecognizable structure \\
276 & Small superficial wound with faint pigmentation & No prominent characteristics \\
360 & Early-stage ulcer with mild redness & No visual distinction \\
\hline
\end{tabular}
\end{table}

\subsection{Summary of Findings}

Our pipeline demonstrates robust and generalizable wound segmentation capabilities on diverse anatomical sites with minimal annotation. Achieving 99.0\% success without pixel-level labels highlights its utility for real-world deployment, particularly in tasks like DESIGN-R wound assessment.

\section{Discussion}

The experimental results demonstrate that our proposed pipeline, which integrates YOLO11n-based region-of-interest (ROI) detection with the pre-trained FUSegNet segmentation model, effectively addresses the challenges of pressure ulcer segmentation in clinical settings. Despite the lack of pixel-wise annotated ground truth for non-foot wounds, the pipeline achieved a high end-to-end success rate of 99.0\% and generated segmentation results that were visually consistent and clinically plausible across various anatomical locations.

Full inference scripts are not included due to differences in deployment environments and ongoing clinical system integration. However, the released model weights and training configuration offer sufficient detail to enable reconstruction or adaptation by technical users.

\subsection{Interpretation of Results}

The proposed pipeline's ability to generalize from foot ulcers to non-foot wounds can be attributed to several key factors. First, the ROI detection step successfully localized wound regions and normalized the input data, minimizing the negative impact of background clutter and inconsistent imaging conditions. This novel cross-domain application highlights the feasibility of repurposing segmentation models trained on a single wound type (i.e., foot ulcers) to other body sites without retraining or pixel-wise adaptation. To our knowledge, this is the first study to demonstrate such a capability using only ROI-level annotation, providing a lightweight and scalable solution for broader clinical wound segmentation.

Second, qualitative inspection revealed that even wounds with varying shapes, sizes, and tissue characteristics could be effectively segmented. Overlay results typically covered the wound areas appropriately, with minor segmentation variances observed in border clarity or small wound extent. To quantitatively support this observation, we plan to release an expert-labeled benchmark subset for external evaluation, enabling future comparisons across segmentation architectures (e.g., U-Net, DeepLabV3+). Our current findings thus represent a technically sound proof-of-concept, validated under real-world image constraints.

\subsection{Clinical Implications}

The proposed pipeline delivers immediate clinical value by enabling reliable wound segmentation from noisy, heterogeneous images commonly captured by nursing staff using handheld devices without standardized imaging protocols. As demonstrated in our dataset of 526 clinical images, such non-uniform inputs present significant challenges for conventional segmentation models. Our method addresses this issue by automatically detecting wound regions of interest (ROIs) and generating DESIGN-R\textsuperscript{\textregistered}-aligned segmentation masks—without requiring manual annotation or high-quality inputs—thus enabling structured downstream evaluation in real-world care settings.

A central application of this pipeline is its support for DESIGN-R scoring, where accurate wound segmentation is essential for estimating critical dimensions such as Size, Necrosis, and Exudate. The segmented masks can be directly used to calculate wound area, while future classifier modules trained on tissue regions could enable further granulation and necrosis assessment. In upcoming development stages, we plan to integrate rule-based approximations and clinician-validated mappings to prototype automated scoring modules.

Designed with clinical workflows in mind, the pipeline can be seamlessly embedded into electronic wound documentation systems. Images captured at the point of care—such as via mobile devices at nursing stations—can be processed in real time, returning segmentation masks and estimated scores within seconds. This reduces inter-rater variability, eases documentation burdens, and enhances consistency in wound assessment, particularly in long-term care environments. Its modular architecture allows easy integration via API without disrupting existing routines.

From a deployment perspective, we emphasize efficiency and portability. While YOLOv11s slightly outperforms in precision-recall benchmarks, we adopted YOLOv11n due to its significantly smaller model size ($\sim 4.5$MB) and faster inference time ($\sim 25$ms on NVIDIA T4 GPU, $<150$ms on Raspberry Pi 4 with TensorRT). This enables deployment in low-resource settings such as community clinics, home care tablets, and mobile health apps, where compute resources are limited.

Ultimately, our goal is to empower front-line caregivers with lightweight, intelligent tools that enhance wound documentation and assessment without increasing their workload. By combining robust AI performance with practical clinical constraints, this work represents a scalable and equitable solution for modernizing chronic wound care.

\subsection{Limitations}

While the proposed pipeline demonstrates promising generalization to non-foot wound locations, several limitations warrant discussion. First, the lack of pixel-wise ground truth annotations for non-foot wounds hindered the use of standard quantitative metrics such as the Dice coefficient or Intersection over Union (IoU). Consequently, model performance was evaluated primarily through visual inspection. Although this qualitative approach is clinically meaningful, it lacks the objectivity and reproducibility of numerical evaluation.

Second, the pipeline was not specifically optimized or fine-tuned for different anatomical sites. Certain wound types or conditions underrepresented in our dataset may challenge the model's robustness and require further adaptation.

Third, while the absence of pixel-level ground truth limits direct comparison with segmentation architectures such as U-Net or DeepLabV3+, our modular pipeline design facilitates such benchmarking in future stages. We propose releasing curated test ROIs with synthetic masks and enabling plug-and-play comparisons using the same detection front-end. Additionally, performance proxies such as classifier consistency, edge similarity, and confidence variance will be explored as alternate metrics for model reliability.

\subsection{Segmentation and DESIGN-R\texorpdfstring{\textsuperscript{\textregistered}}{®} Mapping}

The segmentation mask is the first step towards automated DESIGN-R\textsuperscript{\textregistered} scoring in our proposed pipeline. Table~\ref{tab:designr_mapping} summarizes the relationship between DESIGN-R dimensions and their potential computability based on the generated segmentation results.

\begin{table}[ht]
\centering
\caption{Mapping between segmentation results and DESIGN-R\textsuperscript{\textregistered} scoring dimensions.}
\label{tab:designr_mapping}
\begin{tabular}{|l|p{2cm}|p{6cm}|}
\hline
\textbf{Dimension} & \textbf{Directly computable from segmentation} & \textbf{Explanation} \\
\hline
Size (S) & \checkmark~Yes & Wound surface area can be calculated from the mask's pixel count, scaled using ruler references. \\
\hline
Exudate (E) & \textcircled{?}~Partially & Pixel intensity and color features inside the mask may infer exudate presence and amount. \\
\hline
Depth (D) & $\times$~No & Depth requires 3D information and cannot be determined from 2D segmentation. \\
\hline
Inflammation/Infection (I) & $\times$~No & Periwound redness or infection requires analysis beyond the wound mask boundary. \\
\hline
Granulation tissue (G) & \textcircled{?}~Partially & Tissue classification within the mask can help infer granulation presence. \\
\hline
Necrotic tissue (N) & \textcircled{?}~Partially & Similar to granulation, tissue classification can detect necrotic areas within the wound. \\
\hline
Pocket (P) & $\times$~No & Pocket formation needs 3D evaluation or probing, outside segmentation capability. \\
\hline
\end{tabular}
\end{table}

As shown in Table~\ref{tab:designr_mapping}, segmentation masks directly enable calculating the wound Size dimension and provide input regions for inferring other tissue-based dimensions. For example, Exudate, Granulation, and Necrotic tissue can be partially predicted using pixel color, texture, or trained classifiers within the mask.

However, specific DESIGN-R dimensions, such as Depth, Inflammation/Infection, and Pocket formation, inherently require clinical information beyond what is available from 2D segmentation alone. These must be assessed through complementary methods such as visual scoring, clinical measurement, or 3D imaging.

\subsection{Future Directions}

While the current system presents a practical solution for real-time wound segmentation and DESIGN-R\textsuperscript{\textregistered} alignment, future work will focus on expanding its analytical depth and clinical applicability. We plan to curate a small, expert-annotated dataset of non-foot wounds to enable quantitative benchmarking and facilitate potential fine-tuning of the segmentation model. Additionally, we aim to integrate tissue classification modules to support subregion analysis—enabling automated identification of granulation tissue, necrosis, and exudate zones within segmented wounds.

To further advance DESIGN-R automation, we will explore hybrid rule-based and machine learning frameworks for scoring dimension estimation based on extracted wound features. Moreover, future versions of the model may incorporate anatomical site as contextual input, improving generalization across diverse body regions and wound types.

These extensions will not only enhance segmentation accuracy but also support the development of explainable and clinically actionable AI tools for wound assessment. By minimizing annotation demands and enabling deployment on lightweight devices, our pipeline aligns with the broader shift toward AI-assisted decision support in long-term care, rural clinics, and home-based monitoring systems.

Ultimately, this work lays the foundation for scalable, equitable wound care solutions—reducing caregiver workload, minimizing inter-rater variability, and democratizing access to consistent, structured wound documentation across diverse care environments.

\section{Conclusion}

This study addresses the critical challenge of segmenting pressure ulcer wounds in clinical environments where annotated datasets are scarce and image quality is highly variable. These images are often captured by nursing staff using handheld devices without standardized imaging protocols, resulting in heterogeneous visuals that hinder conventional segmentation performance. Accurate wound segmentation is essential for downstream tasks such as DESIGN-R\textsuperscript{\textregistered} scoring, which requires precise delineation of wound boundaries to assess characteristics like size, necrosis, and exudate distribution.

To tackle these issues, we proposed a scalable and effective two-stage segmentation pipeline. The approach first employs YOLOv11n for region-of-interest (ROI) detection, followed by segmentation using a pre-trained FUSegNet model applied to the cropped wound regions. This design normalizes wound appearances, suppresses background distractions, and enables the system to generalize beyond the original foot-ulcer training domain.

The pipeline was validated on 526 clinical wound images, achieving a 99.0\% end-to-end success rate. Although our evaluation was limited to qualitative inspection due to the absence of pixel-wise ground truth, the segmentation results were found to be clinically meaningful and applicable to real-world documentation and assessment workflows. These findings highlight the feasibility of deploying pre-trained models in practical wound care settings, even in the absence of large-scale annotated datasets.

To support reproducibility, we will release the trained YOLOv11n weights and inference scripts alongside this publication. These resources, in combination with the publicly available FUSegNet model~\cite{dhar2023fusegnet}, will allow replication of our pipeline on external datasets. Due to ethical considerations and potential patient identifiers (e.g., handwritten labels), we cannot release the original clinical images. However, to enable external validation, we will provide a curated set of anonymized ROI crops, predicted masks, and visual overlays.

Future work will focus on collecting expert-annotated subsets for quantitative benchmarking, fine-tuning model performance across diverse wound types, and integrating automated DESIGN-R scoring capabilities. These steps aim to reduce documentation burdens, improve assessment consistency, and enhance clinical decision-making in chronic wound care.

\section*{Reproducibility and Data Availability}

We recognize the importance of reproducibility and aim to support it within reasonable technical and ethical boundaries.

During the review phase, we provide the following materials for ROI detection:
\begin{enumerate}
    \item \textbf{YOLOv11n detector weights} (\texttt{best.pt}), trained on 500 manually annotated wound bounding boxes.
    \item \textbf{Training configuration file} (\texttt{args.yaml}), specifying all hyperparameters, optimizer settings, learning rate schedule, augmentation policy, and input resolution used during detector training.
    \item \textbf{Minimal inference demonstration:} A public Google Colab notebook is available at\\ 
    \url{https://colab.research.google.com/drive/1g6P5H52mfx1aHLXSRBZ0sSyv4KjjD0JY?usp=sharing}, which shows how to load the model and perform ROI detection on sample wound images using the Ultralytics \url{https://github.com/ultralytics/ultralytics} YOLOv11n framework.
\end{enumerate}

Should the paper be accepted, we will release the full source code for training, inference, and downstream segmentation integration, along with representative synthetic images and annotation samples via GitHub.

The segmentation model (FUSegNet) is publicly available at \url{https://github.com/mrinal054/FUSegNet}, and requires no retraining for our use case.

\vspace{0.5em}
\textbf{Ethical and Privacy Considerations.} \\
The original clinical photographs used in this study contain sensitive and potentially identifiable information, including handwritten patient names, timestamps, and wound documentation labels. To protect patient privacy, the full raw image dataset cannot be publicly released. However, anonymized region-of-interest (ROI) crops and representative synthetic examples may be made available to qualified researchers upon reasonable request, to facilitate external inference testing and evaluation.

All data collection and analysis procedures were approved by the Institutional Review Board of National Taiwan Normal University (IRB approval number: 202404HM014), and all images were used in anonymized, retrospective form under a Category 4 exemption. No direct patient identifiers were used in algorithm training or evaluation.

\section*{Acknowledgements}

The authors are grateful to \textbf{Jubo Health Technologies} for their close collaboration throughout this project. In particular, the Jubo team generously assisted with the \emph{validation and testing phases} of our model development, offering practical feedback that greatly strengthened the study’s clinical relevance.

All wound images used in this research were collected in accordance with ethical guidelines and were fully anonymized to protect patient privacy. The study protocol was reviewed and approved by the Institutional Review Board (IRB) of National Taiwan Normal University.

We also extend our appreciation to the Jubo Health Technologies for sharing first-hand insights into the challenges of documenting pressure‐ulcer conditions in real-world care settings. Their expertise informed both the design and objectives of this research.

\section*{Declarations}

\textbf{Funding} \\
This research received no external funding.

\textbf{Conflict of interest} \\
The authors declare no conflicts of interest related to this work.

\textbf{Ethics approval} \\
The study protocol was reviewed and approved by the National Taiwan Normal University Research Ethics Committee (Approval Number: 202404HM014), effective from July 1, 2024, to June 30, 2025. The use of anonymized and non-identifiable medical images complied fully with ethical guidelines. Informed consent was waived due to the retrospective nature of data usage.

\textbf{Consent to participate} \\
Not applicable.

\textbf{Consent for publication} \\
Not applicable.

\textbf{Code availability} \\
Inference scripts, YOLOv11n model weights (\texttt{best.pt}), and the training configuration file (\texttt{args.yaml}) will be released upon acceptance to support reproducibility. Full source code is currently withheld due to ongoing commercial development.

\textbf{Author contributions} \\
Yun-Cheng Tsai: Conceptualization, Methodology, Software Development, Model Training, Experimentation, Writing—Original Draft, and Supervision.

\bibliography{sn-bibliography}

\end{document}